%% file: main.tex
\documentclass[runningheads]{llncs}
\usepackage{graphicx}
\usepackage{cite}
\usepackage{tablefootnote}
\usepackage{booktabs}
\usepackage{hyperref}
\usepackage[utf8]{inputenc}
\usepackage{amssymb}
\usepackage{amsmath}
\usepackage[ruled,vlined]{algorithm2e}
\usepackage{algpseudocode}
\usepackage[compatibility=false]{caption}

\begin{document}
\title{Efficient Training Convolutional Neural Networks on Edge Devices with Gradient-pruned Sign-symmetric Feedback Alignment}


%
%
\author{Ziyang Hong \orcidID{0000-0001-9096-1750} \and
C. Patrick Yue \orcidID{0000-0002-0211-2394}}
\authorrunning{Ziyang Hong \and C. Patrick Yue}
\titlerunning{EfficientGrad: Gradient-pruned Sign-symmetric Feedback Alignment}
%
\institute{HKUST-Qualcomm Joint Innovation and Research Laboratory\\
            Dept. ECE, Hong Kong University of Science and Technology \\
\email{zhongad@connect.ust.hk}
\email{eepatrick@ust.hk}}

\maketitle              
\begin{abstract}
With the prosperity of mobile devices, the distributed learning approach enabling model training with decentralized data has attracted wide research. However, the lack of training capability for edge devices significantly limits the energy efficiency of distributed learning in real life. This paper describes a novel approach of training DNNs exploiting the redundancy and the weight asymmetry potential of conventional back propagation. We demonstrate that with negligible classification accuracy loss, the proposed approach outperforms the prior arts by 5x in terms of energy efficiency. 

\end{abstract}
\input{introduction}

\input{related}
\input{generic}
\input{efficientGrad}

\input{evalutation}

\section*{Acknowledgement}
This work is supported by the Hong Kong Research Grants Council under General Research Fund (GRF) project no. 16215620 and the HKUST-Qualcomm Joint Innovation and Research Laboratory.

\bibliographystyle{splncs04}
\bibliography{main.bib}





\end{document}

%% file: introduction.tex
\section{Introduction} \label{Introduction}
\begin{figure}[!ht]
\includegraphics[width=\textwidth]{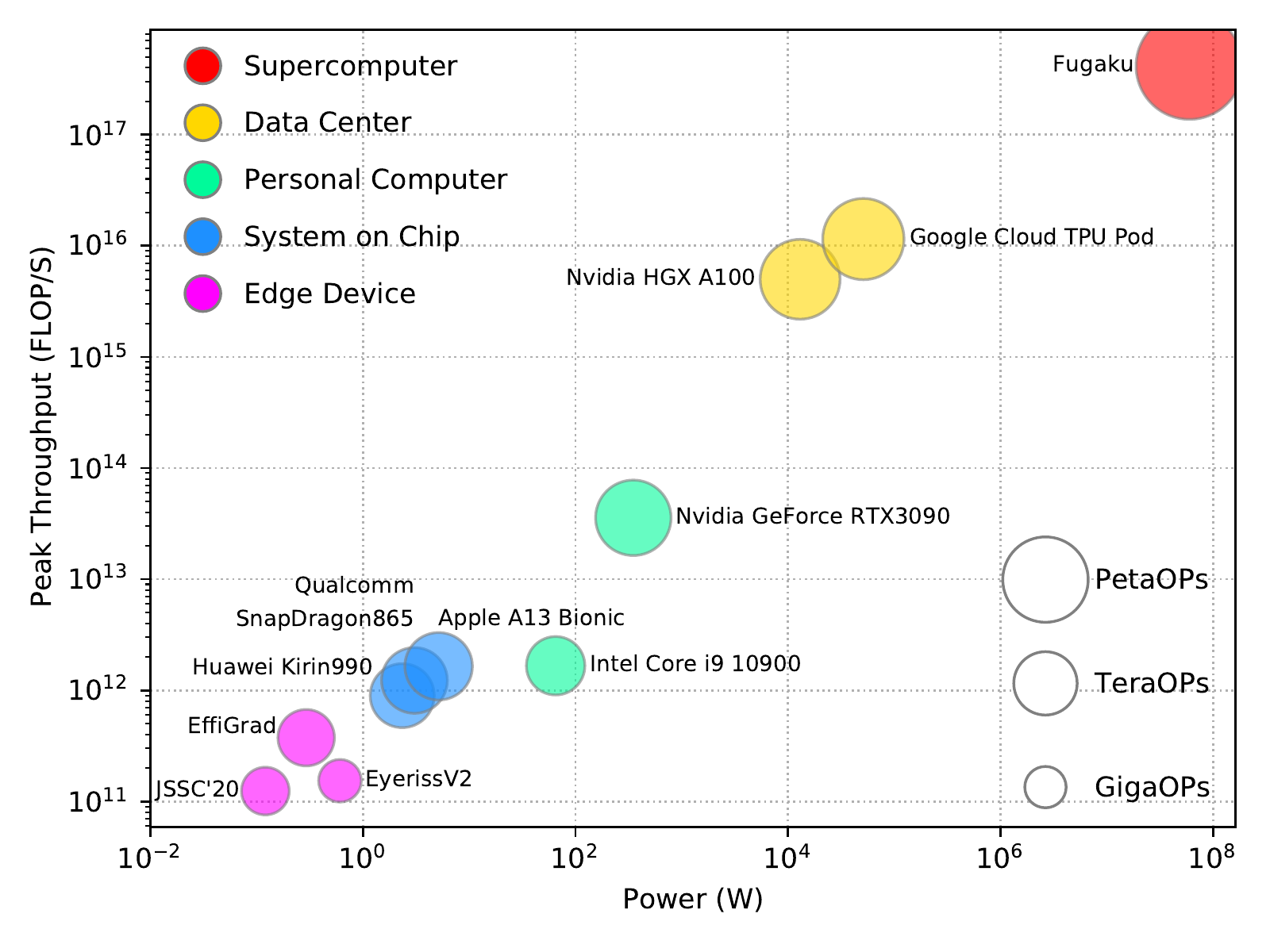}
\caption{Throughput vs. power comparison of the hardware hierarchy.} \label{throughput_power}
\end{figure}

The rapid development of deep learning (DL) algorithms brings the Artificial Intelligence (AI) within the bounds of possibility. 
The prosperity of deep neural networks (DNNs) applications requires a huge amount of feature data. However, there are limitations of centralizing the data for model training. Such huge amount of local data makes it harder to upload due to the limitation of communication bandwidth. Also, people tend to be conservative more and more when it comes to sharing their private data. For instance, for the facial recognition, the data cannot be uploaded to the cloud for training purpose. These issues give rise to the federated learning \cite{YangQiang2019}, which investigates the collaborative model training and inference attacks and constructs robust DNNs models. In the federated learning, rather than the local data, the client nodes such as edge devices always send the updated models or gradients to the central server, which requires the client nodes to be equipped with the capability of model training/re-training. Most of the studies\cite{towards_federated, YangQiang2019, Communication_eff_decentralized} on federated learning work on the encryption and model sharing strategy primarily while the lack of local model re-training capability for underlying hardware remains an issue. With current model training approaches, both the throughput and the power offered by edge devices are not enough for federated learning scenario, due to its constraints of power and inferior energy efficiency shown in the Fig. \ref{throughput_power}.


What is worse, as the computational complexity of the myriad DNNs is increasing, the demand for higher throughput rises drastically. Besides, the slow-down of the Moore's Law makes it harder to meet such a requirement solely depending on technology scaling. The pressure to improve energy efficiency by elaborately designing the specific accelerator architecture on architects and circuit designers becomes higher.

One of the significant factors of limiting the energy efficiency of the accelerators is the excessive external memory access. Containing many epochs until convergence, the model training process of DNNs consumes a great amount of energy by accessing off-chip DRAM for each training sample. Horowitz\cite{Horowitz2014} shows the energy consumption of basic arithmetic and memory operations in a 45nm CMOS process is dominated by the DRAM access, which is more than 200x larger the average of other operations. 
In such case, the power consumption solely brought by DRAM access is 
always beyond the power envelope of typical edge devices such as mobile phones, even in inference process. It poses great challenges on the edge devices for training tasks. 

With these considerations in mind, the aim of this work is to design an efficient training algorithm that leverages the redundancy of conventional model training such that the energy efficiency of the accelerators during training is improved. We propose a novel approach called EfficientGrad, which can reduce DRAM access while maintaining high throughput by utilizing the sign-symmetric feedback. As we discussed in more detail in Section~\ref{Related Work}, the symmetric modulatory signal used in conventional back-propagation-based training algorithm is replaced by the sign-symmetric feedback, for both convolutional layers and fully-connected layers. To eliminate the overhead brought by minor gradients calculation in the backward phase while preserving the original training accuracy, the error gradients generated by sign-symmetric feedback is further pruned in a stochastic fashion.
The main contributions of the paper are summarized as follows:
\let\labelitemi\labelitemii
\begin{itemize}
  \item \textbf{EfficientGrad algorithm} which imposes the sign-symmetric fixed feedback as the modulatory signals for error gradients and prunes the error gradients that follow in a stochastic approach. To ease the classification accuracy drop of CNNs, the stochastic error gradient pruning is described to maintain the learning capability by reaching low angles between the modulatory signals prescribed by itself and back propagation.
  \item \textbf{Specific data reused architecture} to utilize the sparsity and memory access reduction brought by EfficientGrad. By eliminating the transposed weight matrix fetching/storing and minor gradients being involved in backward phase, the energy efficiency is dramatically increased.
\end{itemize}

Our PyTorch \cite{pytorch} implementation used to train the benchmark models with gradient-pruned sign-symmetric feedback alignment is available at \url{https://github.com/HaFred/EfficientGrad}.

%% file: related.tex
\section{Related Work} \label{Related Work}
There has been substantial studies focusing on accelerating the inference of popular DNNs \cite{Eyerissv2, EIE}, but no too much efforts are invested on training process for dedicated edge devices. Even though DaDianNao \cite{DaDianNao} proposed a prototype accelerator for convolutional training, the power consumption is with 14W therefore it is too large for edge devices, which is normally around hundreds of mW as shown in the Fig. \ref{throughput_power}. \cite{BFA} proposes a low-power training processor utilizing feedback alignment, which was first presented by \cite{feedback_nature}, to reduce the energy consumption brought by external DRAM access . However, \cite{BFA} limits the training capability within fully-connected layers by disabling convolutional training. This is because as \cite{feedback_nature} analyzes, the angle of the error gradients between back propagation and feedback alignment will be stuck with $90^\circ$ therefore no learning happens. To deal with this problem, \cite{HowWeight, DRTP, feedback_nips} work on the variants of feedback alignment. In section. \ref{EfficientGrad}, we show our design solve the aforementioned issues and achieve the learning capability on convolutional layers while minimizing the computations needed in the training process. 

%% file: generic.tex
\section{Generic Back Propagation}
\begin{figure}[ht]
\includegraphics[width=\textwidth]{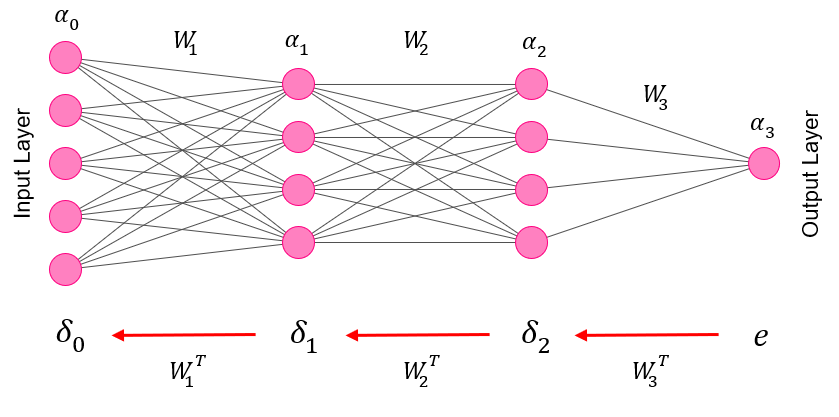}
\caption{The forward phase and backward phase in back propagation.} \label{bp}
\end{figure}

\begin{algorithm}[H]
\SetAlgoLined
\KwIn{[$Img_{1}$, $Img_{2}$, ..., $Img_{N}$]: Input batch with the size of $N$ images, [$W_{1}$, $W_{2}$, ..., $W_{L}$]: $L$ layers of trainable weights}
\KwOut{Trained network for inference}
\CommentSty{// Phase 1: Forward}

\For {$l \leftarrow 0$ \KwTo $L-1$}{
    $a_{l+1} = \sigma(W_{l+1}\ast a_{l})$ \;
    \If{$l=L-1$}{$Loss=C(a_{l+1}, y)$}
}
\CommentSty{// Phase 2: Backward}

\For {$l \leftarrow L$ \KwTo $1$}{
    \eIf{$l=L$}{$e=\frac{\partial Loss}{\partial a_{l}}=
    (a_{l}-y)\odot\sigma'(a_{l})$}{$\delta_{l}=W^{T}_{l+1} \ast \delta_{l+1} \odot \sigma'(a_{l})$}
    }
\CommentSty{// Phase 3: Weight gradients update}

\For {$l \leftarrow L$ \KwTo $1$}{
    $\Delta W_{l}=\frac{\partial Loss}{\partial W_{l}}=a_{l-1} \ast \delta_{l}$\;
    $W_{l} = SGD(W_{l}, \Delta W_{l}, lr=\gamma, momentum=\mu)$
}
 \caption{Generic Convolutional Neural Networks Training}
 \label{algo1}
\end{algorithm}

Back Propagation (BP) algorithm \cite{backprop} and the Stochastic Gradient Descent (SGD) algorithm \cite{SGD} are the canonical approaches for DNNs training. 
They remains powerful and effective and being used invarious nowadays AI systems. 
Generally, the BP algorithm is an efficient use of the Chain Rule for generating gradients, and the SGD algorithm takes the average of gradients of a mini-batch input to update the weights. 
The conventional way of performing back propagation is illustrated in Fig. \ref{bp}. 


In Algo. \ref{algo1}, a given learning rate $\gamma$ will be used as the coefficient of the weight gradients. Additionally, the previous updating value is also taken into account by a factor of the momentum $\mu$ to help accelerate SGD in the target direction and mitigates oscillations. 
$N$ denotes the batch size of each training iteration. Practically, we will neither take the whole training data set as Batch Gradient Descent (BGD) nor a single training sample for every weight update. Consequently, mini-batch SGD,  which updates parameters by randomly select a mini-batch $N$ of training sample from the training set, is utilized. It helps the optimization to jump out of local minimum comparing to BGD. The training algorithms for CNNs consists of three phases as in Algo. \ref{algo1}. 

%% file: efficientGrad.tex
\section{EfficientGrad} \label{EfficientGrad}
\subsection{Algorithm}
Consider the feedback alignment proposed by \cite{feedback_nature}, the modulatory signals in the phase 2 of Algo. \ref{algo1} is a random feedback matrix $B$, which eliminates the usage of transposed or 180 degree rotated weight matrix $W^{T}$ as shown in Fig. \ref{bp}:
\begin{equation}
\label{vani_fa}
    \delta_{l} = B_{l+1} \ast \delta_{l+1} \odot \sigma'(a_{l})
\end{equation}


Nevertheless, as mentioned in \cite{feedback_nature}, the defect of feedback alignment is that the fixed feedback cannot be directly imposed on convolutional layers. It is because all the neurons with in a convolutional layer share precisely the same receptive field, and such weight sharing aggravates the regularization effect of feedback alignment and leads to over-regularization. Beyond that, the activation function $\sigma$ in \cite{feedback_nature} compromises into hyperbolic tangent. From our experiments, We observe that in the early training stages, the regularization effect of feedback alignment will often improperly impel the activation into negative region, which will lead to dead neurons if ReLU is applied.

To address these limitation for feedback alignment on CNNs, we mitigate the over-regularization issues by assigning the fixed random feedback with the symmetric signs of its corresponding weights. Moreover, to restore the improper killed neurons in the hidden layers, we append batch normalization \cite{batchnorm} layers in between wherever the neurons tend to be killed. The sign-symmetric feedback alignment in the phase 2 of Algo. \ref{algo1} could be obtained:
\begin{equation}
\label{ssym_fa}
    \delta_{l} = sign(W_{l+1}) \odot \mathopen| B_{l+1} \mathclose| \ast \delta_{l+1} \odot \sigma'(a_{l})
\end{equation}

\begin{figure}[ht]
\includegraphics[width=\textwidth]{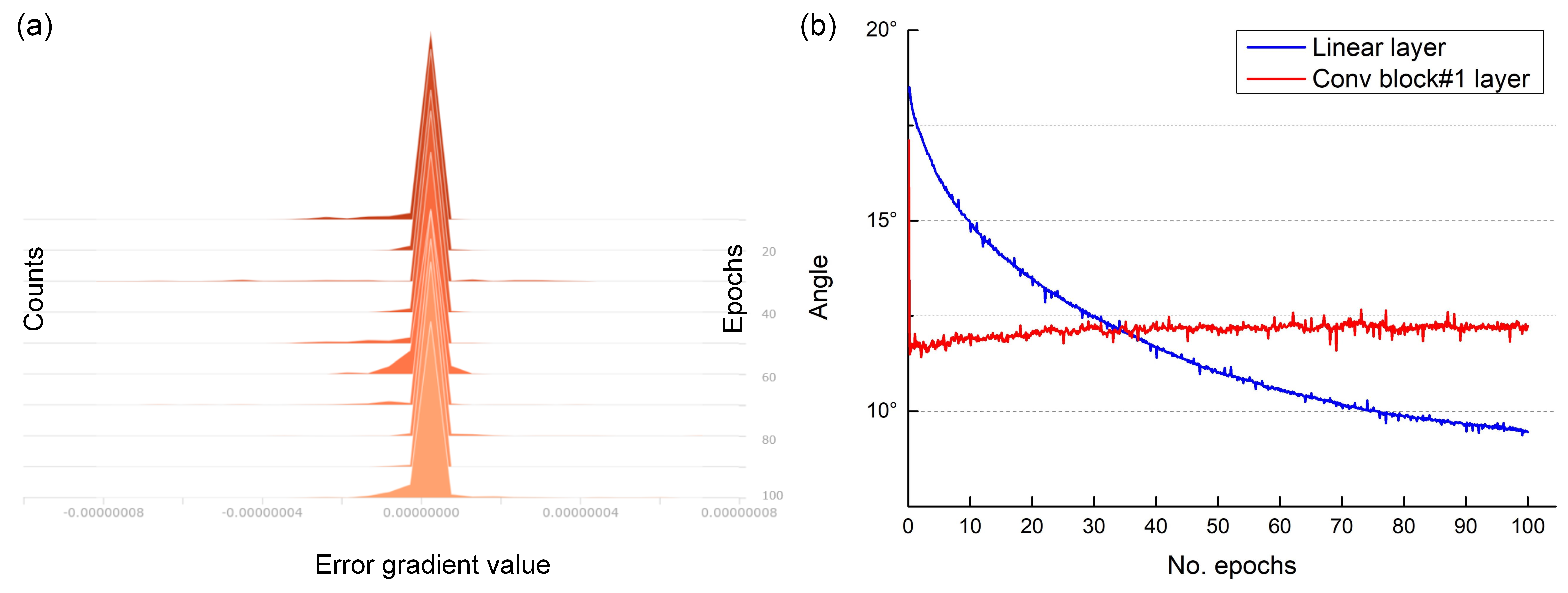}
\caption{(a) Error gradients $\delta$ distribution of ResNet-18 over 100 epochs, (b) The angles between error gradients prescribed by BP and those prescribed by EfficientGrad of two representative layers.} 
\label{res_dist_angle}
\end{figure}

Besides, the resulting error gradients in (\ref{ssym_fa}) turns out to be small in magnitude. We observe that the error gradients of adopting sign-symmetric feedback alignment in the phase 2 of Algo. \ref{algo1} is distributed in a long tailed normal distribution, which is shown in the Fig. \ref{res_dist_angle}(a). It means that the computation brought by (\ref{ssym_fa}) can be abandoned as long as its expectation remains. Inspired by \cite{accele_prune}, we propose a stochastic gradient pruning algorithm on (\ref{ssym_fa}) to reduce these redundant gradient computations. The basic idea is to prune the error gradients prescribed by sign-symmetric feedback while maintaining their mathematical expectation. To make expectation remain unchanged, it is natural to compensate the pruned gradients back to the pruned threshold:

\begin{equation}
\label{stochastic_tau}
      \hat{\delta_{l_{i}}} =
    \begin{cases}
        \delta_{l_{i}} & \text {if $\mathopen| \delta_{l_{i}}\mathclose| > \tau$}\\
      \tau\odot sign(\delta_{l_{i}}) & \text{if $\tau \geq \mathopen| \delta_{l_{i}}\mathclose| \geq r\tau$}\\
      0 & \text{otherwise}
    \end{cases},     r\in[0, 1]
\end{equation}
where $r$ is a uniform random number ranging from 0 to 1. Note that (\ref{stochastic_tau}) is applied on top of (\ref{ssym_fa}), we need to ensure that the angles of error gradients with EfficientGrad is well under $90^\circ$. Since the error gradients are pruned with expectation remains, the sign-symmetric feedback stays unchanged, thus it is still affecting the weight to be  aligned with the random fixed feedback as analyzed in \cite{feedback_nature}. Comparing to the original feedback alignment, the sign-symmetric one with stochastic gradient pruning could be even reach lower angle under $45^\circ$, the angles over 100 epochs training on ResNet-18 \cite{ResNet2016} is shown in the Fig. \ref{res_dist_angle}. As discussed in Section.\ref{Related Work}, the lower angle between error gradients the better learning capability. The linear (fully-connected) classifier layers keeps align with the random feedback because of the over-regularization is suppressed in fully-connected layers, whereas the convolutional layers drops fast but tend to be stable. It makes sense because the batch normalization layer liberates the neuron turn-off problems mentioned above and restores the internal covariate shift layer-wisely.

One of the critical parts of EfficientGrad algorithm is to determine a dynamic pruning threshold $\tau$ that will preserve the original accuracy that a given DNNs model could reach. Consider the cumulative density function (CDF) $\Phi$ of a given $\delta_l$, if we use a pruning rate $P$ to control the gradient sparsity, then (\ref{tau_deter}) holds.
\begin{equation}
\label{tau_deter}
    P=1-\{[1-\Phi(\frac{\tau}{\sigma})]\times2\}=2\Phi(\tau)-1
\end{equation}
\begin{equation}
\label{tau_deter2}
    \tau=\Phi^{-1}(\frac{1+P}{2})\cdot\sigma
\end{equation}
With (\ref{tau_deter2}), the ratio of $\delta_l$ which will be stochastic pruned in  (\ref{stochastic_tau}) is set as P. The expectation of $\delta_l$ in EfficientGrad is almost unchanged, thus it leads to a negligible classification accuracy loss.





\subsection{Hardware}

\begin{figure}[ht!]
\includegraphics[width=\textwidth]{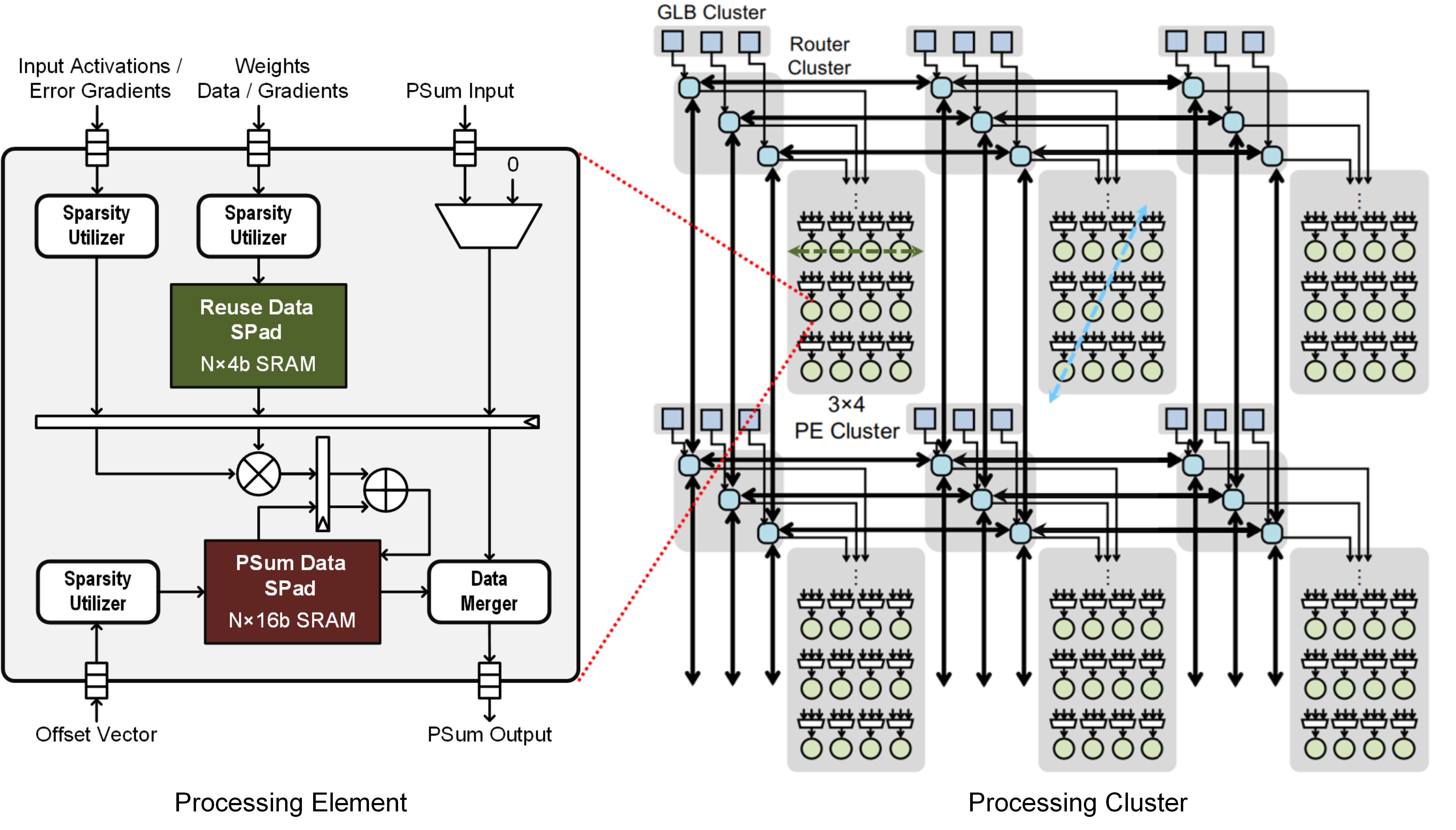}
\caption{Overall architecture of the proposed EfficientGrad DNNs Accelerator.} 
\label{hardware}
\end{figure}

To enhance the energy efficiency of edge devices for DNNs training task, we design the architecture that leverages the sparsity and memory access reduction brought by EfficientGrad. The DNNs mapping dataflow is based on the row stationary proposed in EyerissV2 \cite{Eyerissv2}. 
Our DNNs training accelerator (in Fig. \ref{hardware}) consists of 6 Processing Clusters (PCs). Each of the cluster is made up of 12 Processing Elements (PEs) which is in charge of multiply-accumulations (MACs). Within a PC, the weight matrix row is preloaded and shared across each row of the PC. That is to maximize the convolutional kernel reuse especially when the activation size gets larger. Besides, the activation row is shared anti-diagonally across each PC. In such case, the activation row needs to be operated by certain weight row is manipulated in a systolic array fashion without stall. There is a router cluster and global buffer cluster (GLB) for each PC to communicate PC-wisely and storing the data that is yet to push back to the external DRAM.

Unlike EyerissV2 that only supports CNNs inference, each PE of EfficientGrad accelerator can proceed all three phases of training in Algo. \ref{algo1}. The weight in phase 1 and fixed feedback in phase 2 are both stored in the reuse data scratchpad inside the PE. It maximizes the convolutional reuse for all three phases and minimizes the external memory access in phase 2.

%% file: evalutation.tex
\section{Evaluation} \label{Evalutation}

\begin{figure}[h]
\includegraphics[width=\textwidth]{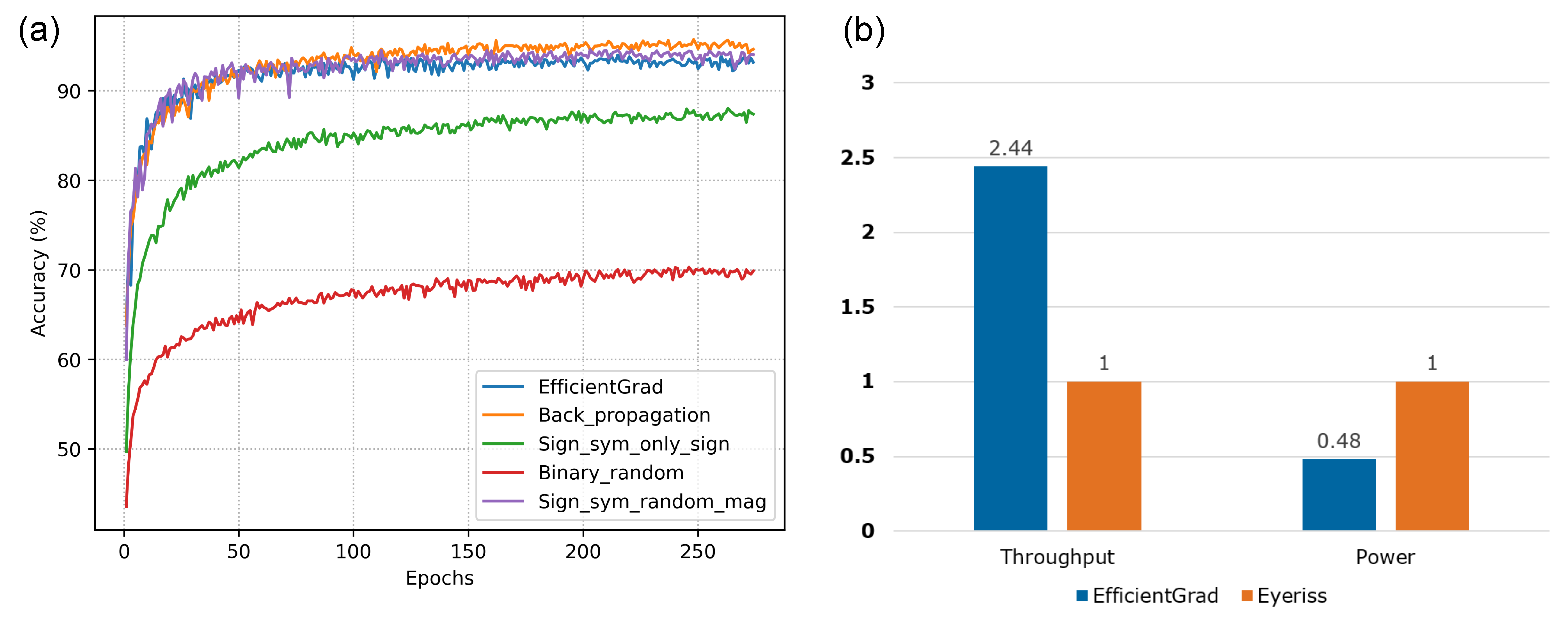}
\caption{(a) Classification accuracy convergence of ResNet-18 for training over 270 epochs, (b) The performance of our proposed EffcientGrad comparing to EyerissV2.} 
\label{evaluation_result}
\end{figure}

We compare the classification accuracy of ResNet-18 on CIFAR-10\cite{cifar10} of EfficientGrad with other feedback alignment variants \cite{HowWeight, BFA} in Fig. \ref{evaluation_result}(a). For deep CNNs training, binary random feedback in \cite{BFA} and sign-symmetric only feedback in \cite{HowWeight} degrade in terms of accuracy. EfficientGrad compromises negligible accuracy loss over sign-symmetric random magnitude feedback, to achieve less backward phase calculation.

The hardware part of EfficientGrad is designed in Chisel\cite{chisel} with the behavioral simulation with ChiselTest\cite{chiselTest} and synthesized using the Synopsys Design Compiler with SMIC 14nm tt process. Utilizing the fork-join property of ChiselTest, we are able to build a simulation-based timing model in scala, to handle the intra-PE pipeline. The power and clock rate of the SRAM/RegFile is generated by the 14nm off-the-shelf memory compiler. EfficientGrad can achieve peak throughput at 121GOP/S with the clock frequency of 500MHz and the power of 790mW theoretically. In such case, for one patch forward phase of training on ResNet-18, EfficientGrad can finish within 0.69ms. 

We normalize the throughput of EfficientGrad with reference to the unpruned back propagation version of EyerissV2\cite{Eyerissv2}. It is significantly faster than EyerissV2 with the 2.44x throughput improvement and 0.48x power reduction as shown in the Fig. \ref{evaluation_result}(b). We also include EfficientGrad in Fig. \ref{throughput_power} to compare with other popular computational devices. It appears that EfficientGrad reaches higher energy efficiency and is well-suited to edge devices.

\section{Conclusion}
In this paper, we present EfficientGrad, an efficient back-propagation-based DNNs training algorithm that enables one to make full use of both the elasticity of weight symmetry problem, and the redundancy residing in the conventional back propagation algorithm. As demonstrated in the paper, our proposed EfficientGrad can increase the throughput by approximately 2.5x and decrease the power consumption to a half, which leads to superior energy efficiency as 5x against prior accelerators.